\documentclass[letterpaper, 10 pt, conference]{ieeeconf}  

\IEEEoverridecommandlockouts                              

\usepackage{graphicx}
\usepackage{tabularx}
\usepackage{url}
\usepackage{breakurl}

\title{\LARGE \bf
Virtual Reality as a Tool for Studying Diversity and Inclusion in Human-Robot Interaction: Advantages and Challenges
}

\author{André Helgert$^{1}$, Sabrina C. Eimler$^{1}$, Carolin Straßmann$^{1}$ 
\thanks{$^{1}$Computer Science Institute, University of Applied Sciences Ruhr West,
Bottrop, Germany; \{{andre.helgert, sabrina.eimler, carolin.strassmann}\}@hs-ruhrwest.de;
        }%
}

\begin{document}

\maketitle
\thispagestyle{empty}
\pagestyle{empty}

\begin{abstract}
This paper investigates the potential of Virtual Reality (VR) as a research tool for studying diversity and inclusion characteristics in the context of human-robot interactions (HRI). Some exclusive advantages of using VR in HRI are discussed, such as a controllable environment, the possibility to manipulate the variables related to the robot and the human-robot interaction, flexibility in the design of the robot and the environment, and advanced measurement methods related e.g. to eye tracking and physiological data. At the same time, the challenges of researching diversity and inclusion in HRI are described, especially in accessibility, cyber sickness and bias when developing VR-environments. Furthermore, solutions to these challenges are being discussed to fully harness the benefits of VR for the studying of diversity and inclusion.

\end{abstract}

\section{Introduction}


Virtual Reality (VR) has established itself as a powerful research tool in human-computer interaction through the use of unique capabilities and immersive experiences \cite{Martingano.2021}. This is not only due to the fact that VR can be used to create dynamic and controllable environments, but also because experimental studies have shown that VR-based assessments are just as effective compared to traditional surveys and represent human behavior and physiological patterns just as they do in real-world scenarios \cite{Roberts.2019}. For this reason, VR is also increasingly used as a research tool for human-robot interactions (HRI) \cite{Arntz.2021, Liu.2017}. 
Since communication between humans and robots can be very different and individual due to various diversity characteristics, it is important that robots can respond to them appropriately. As society becomes more diverse, diversity-related challenges arise that increasingly affect us in the private and public places. This development is not only challenging in human-human interaction, but also affects all computer systems with which humans interact directly, including robots. Therefore, robotic systems must be able to respond to and interact with each individual, taking into account individual diversity factors. In this paper, we will discuss the capabilities of VR in the study of HRI, especially in terms of diversity and inclusion research, with advantages in the field of controllability, manipulability, flexibility and extended measuring methods as well as its challenges and possible solutions.

\section{VR as a Research Tool}

VR is an important subset of immersive technologies that is increasingly being used in HCI and psychological studies. One reason is the possibility to conduct studies which are difficult to perform or control in the real world. Thus, in today's world, VR appears as one of the most beneficial tools to achieve effective results in the field of therapy and rehabilitation in patients \cite{Rizzo.2004}, such as in anxiety disorders \cite{MaplesKeller.2017} or in developing skills to deal with pain \cite{Chan.2018}. Furthermore, the use of VR in the fields of educational \cite{Parong.2018}, social \cite{Carlo.2002} and experimental psychology \cite{Vasser.2015}, among others, has produced advantages in data collection methods and measurement data that are difficult to implement in real-world scenarios. 

VR is also already being used extensively in HRI, providing researchers with new ways to explore and understand the complexity of human-robot interaction. For example, Shariati et al. designed a VR-robot based on a real social robot, which was specifically designed to improve learning, clinical therapy for children with chronic diseases, and education \cite{Shariati.2018}. The virtual robot was compared with a real robot in an experimental study and the results indicate that the acceptance of a VR-robot is the same as that of a real robot, and that the virtual robot did not present a significant difference in performance from the real robot. Shariati et al. conclude that the VR-platform has the potential to have an important auxiliary solution for social robot research. Another controlled study was conducted to investigate whether VR can be a suitable platform for exploring social interactions between humans and robots \cite{Sadka.2020}. The quantitative results suggest that the core aspects of human-robot social interactions are preserved in a VR-simulation. These examples show that the medium VR is already used in HRI to produce valid research results. It is therefore natural to explore VR in relation to diversity and inclusion in HRI. Common advantages of VR applications can be applied to the diversity topic, which will be explained in detail in the next chapter.

\section{Advantages in Exploring Diversity and Inclusion in VR}

In addition to the countless applications opportunities of VR, it can also be used to explore diversity and inclusion by providing exclusive characteristics of VR, but also by extending common measurement methods. Since VR is used for a wide variety of research purposes, the following characteristics are presented in the context of HRI-studies.

\subsection{Controllable Environment}

The use of VR as a research tool in HRI provides a significant advantage by allowing studies to be conducted that are difficult to control in the real world \cite{Villani.2018, Todorov.2008}. Conducting studies in a controlled virtual environment helps minimize or even eliminate confounding factors that can occur in field studies. This has the advantage that research results can be generated without the influence of external factors. In addition, every variable can be controlled, from visual to auditory stimuli and even haptic impressions. This level of controllability allows researchers to simulate and manipulate a variety of scenarios to investigate specific questions. Specifically in relation to diversity and inclusion, for example, social interactions can be controlled in their nature of communications and adaptivity to diversity factors. Investigations can be limited to specific situations, which may not be as easy to conduct due to a non-controllable environment. This circumstance ensures a high degree of standardization and reproducibility, which are crucial aspects of controlled studies.  

\subsection{Manipulability}

A key advantage of using VR in HRI research is the high degree of manipulability it offers. Unlike physical robots, VR-simulations can encompass a wide range of robot variations, allowing researchers to explore various factors such as robot appearance, behavior, and capabilities \cite{Negi.2008}. Researchers can manipulate these variables to study how different robot attributes affect human perception, attitude, and behavior. For example, they can study how variations in the robot's gender, ethnicity, or voice affect users' trust, engagement, and willingness to interact. Of course, this manipulability can also be used to study diversity and inclusion characteristics, as different people have different expectations and perceptions based on their background, prior knowledge and so on. However, in addition to manipulating the robot, the environment itself can also be dynamically adapted. Thus, the robot and the interaction with humans can be tested and evaluated in different scenarios and use cases. This allows possible conclusions to be drawn about a wide variety of variables in the environment and an interaction with respect to the perception of the HRI.

\subsection{Flexiblity}

VR-environments provide researchers with the flexibility to tailor interactions e.g. based on specific diversity factors \cite{Bricken.1991} which are difficult or impossible to implement in the real world. By integrating features such as language preferences, cultural backgrounds, or even physical impairments, researchers can create personalized experiences for participants. This customization promotes inclusivity and enables the examination of human-robot interactions across various demographic groups. This may help to counteract discriminatory bias and minimize the reproduction of stereotypes by the robot. In the area of accessibility, the use of VR offers a wide range of possibilities, which can be explored and tested. Features related to accessibility could be, for example, speech or gesture recognition or adaptive interfaces for people with physical disabilities \cite{Bhuiyan.2011}. Understanding how different individuals or communities interact with robots can inform the design of more inclusive and culturally sensitive robotic systems. For example, needs of people with physical limitations or other physical characteristics such as different body sizes can be captured by simulating these situations on their own game character in VR. Through these simple technical adjustments, different diversity characteristics and their needs can be identified, which could provide for a more inclusive experience with the robot in future iterations.




\subsection{Extended Measuring Methods}
VR is commonly used as a research method to study human behavior and cognition because, when properly utilized, it provides an expanded data foundation through specialized measurement methods \cite{Martingano.2021}. There are several measurement methods for researching human-robot interaction in VR. Common methods include eye tracking \cite{Pfeiffer.2013} and physiological measurements \cite{Gaggioli.2014}. For example, eye-tracking technology can be used to collect data on a person's eye movements and what they are looking at and for how long. From this, behavior patterns and focal points of attention can be identified. Physiological surveys could unobtrusively collect data \cite{Betella.2014} on the person's emotional state or measure stress in interaction with the robot and cognitive load. With regard to diversity factors, conclusions could be drawn from people with different characteristics on the perception of the robot and the environment, which can subsequently help to make the human-robot interaction more diversity-friendly. Both methods can be an extension of conventional measurement methods and reveal implications for HRI.

\section{Challenges in Exploring Diversity and Inclusion in VR}

In addition to the advantages of VR, which can represent a more diversified exploration and design of different scenarios, the technology can also have an exclusive or even discriminatory effect. In the following sub-chapters, these challenges are discussed, in particular the accessibility problems, the cyber sickness phenomenon and the bias that developers bring to VR-systems.

\subsection{Accessibility}
The VR-technology may not be accessible for all individuals, especially for people with disabilities. Since conventional VR-devices consist of a headset and two controllers, the use may work better or worse for different limitations. For example, visual impairments affect the perception of the virtual world and, depending on their severity, may be a criterion for exclusion from VR. Hearing impairments can quickly make the virtual world more difficult to use if the design is not inclusive \cite{Jain.2021}. Furthermore, limitations in the mobility of the person can be a problem, since common VR-applications provide for two controllers to move around in the virtual world \cite{Mott.2020}. 

\subsection{Cyber Sickness}

Cyber Sickness (CS) is a syndrome that can occur when using VR, when there is a discrepancy between the visual information projected through VR-goggles and the sensory information perceived by the body. This can lead to nausea and dizziness and affects the VR-experience to a great extent. CS is considered a major barrier to the acceptance and adoption of VR, as there is not yet a fully comprehensive solution for the non-occurrence of the syndrome \cite{Duzmanska.2018}. There are also gender differences in the occurrence of CS. For example, women are more likely to experience cyber sickness when using VR-applications because they have a different average IPD (the distance between the pupils of both eyes) than men \cite{Fledelius.1986} and they cannot adjust their IPD on VR-goggles because they only support the mean IPD of men \cite{Fulvio.2018}.

\subsection{Bias in VR-environments}

Researching diversity and inclusion issues can be challenging because, for example, prejudiced perceptions of developers and researchers and a non-sensitive perception on other social groups can influence the VR-environment itself. These prejudices could possibly be (also less) obviously reflected in the development and design process \cite{Nadeem.2022}. Not only can this be potentially discriminatory, but it can also stigmatize and exclude different social groups. Algorithms based on artificial intelligence (e.g. robotic systems), which are needed for a huge amount of projects in VR can be discriminatory towards certain social groups. For example, a hiring algorithm from Amazon systematically downgraded applications and resumes of female applicants based on biased training data \cite{JeffreyDastin.}. The training set was primarily male centered. Similar discriminatory occurrences exist, for example, in facial recognition \cite{pmlr-v81-buolamwini18a} and in social communication with people \cite{ElleHunt.2016}.

\section{Discussion}

VR offers various ways for researching diversity and inclusion in different settings and szenarios, including HRI. Although we are provided with many new and effective ways to recognize and implement diversity factors, the medium itself can discriminate and could have an exclusive effect. As already discussed, accessiblity, cyber sickness and a bias in VR-environments are a few relevant variables. The question is how we can overcome these challenges in order to exploit the opportunities of VR in terms of diversity and inclusion-friendly implementations. 

When looking at accessibility challenges, there is a very diverse and broad issue. Restrictions on the person are as individual as the person himself. However, special implementations can help to make VR-applications inclusive. For example, there are already solutions in the field of vision impairments, such as those of Zhao et al. who have developed a VR-toolkit, with which various low vision limitations can be improved by the use of e.g. edge measurement and depth enhancement \cite{Zhao.2019}. For hearing impairments, the design of the VR-environments could be adapted to have possible implemented speech outputs put onto the user's glasses via text. For limited mobility or motor impairments it would be possible to use an adaptive controller as a replacement for the two conventional controllers. An adaptive controller such as the Xbox Adaptive Controller by Microsoft \cite{Godineau} can be adjusted and individualized to various constraints. This allows people who are unable to operate a conventional game controller to explore VR-environments and interact with them. Only the virtual representation of the VR-controllers and thus the hand tracking are not (yet) possible. However, this limitation can be minimized by an inclusive design of the VR-environment.

Cyber sickness can be mitigated by ensuring a minimal latency between physical movement in the real world and the corresponding virtual movement in the VR-world. Additionally, optimizing movement settings to avoid abrupt movements and make them feel more natural, like in real life, can help. For example, a jerky motion when climbing stairs in unoptimized VR-applications. Approaches to reduce cyber sickness are currently being investigated. Tian et al., for instance, examined the seating position and whether the virtual coordinate system contradicts the received real-world coordinates through our vestibular system \cite{Tian.2020}. Promising results were achieved in reducing cyber sickness when the real vertical axis aligned with the virtual one.

As many of the previous challenges can be attributed to a certain bias among developers, it is also important to generate a certain awareness among developers. This could not only address existing challenges in the area of diversity and inclusion but also avoid bias in future systems. For this reason, it is important to ensure a certain sensitivity to diversity factors among researchers and developers, but also to constantly question one's own knowledge with regard to diversity.

\section{Conclusion}

VR has established itself as an effective research tool in the field of human-computer interaction. In particular, VR shows its unique capabilities in human-robot interaction research focusing on diversity and inclusion. Through controllable environments, flexibility in customization of interactions and manipulability of robot attributes, VR enables detailed investigation of the complex interactions between humans and robots. Another advantage of VR lies in the advanced measurement methods it offers. Eye-tracking and physiological measurements can be used to capture behavioral patterns and cognitive processes of humans during HRI studies. This allows for analysis of individual perceptions and reactions to the robot and the environment. This helps to make human-robot interactions more diverse and inclusive.

Nevertheless, there are also challenges in exploring diversity and inclusion in HRI with VR. Biases and prejudices of developers and researchers can negatively impact the design and development process and lead to discrimination and exclusion of certain social groups. Therefore, it is important for researchers and developers to be sensitive to diversity factors and continuously challenge their steps regarding diversity. Furthermore, the topics of accessibility and cyber sickness are still challenges today, where there are already initial implications or solutions, which nevertheless represent a hurdle.
Addressing these challenges and promoting diversity and inclusion in VR-based HRI research can lead to more equitable and unbiased human-robot interactions. By harnessing the potential of VR, researchers can help develop robotic systems that address individual diversity factors, promote inclusion, and minimize discriminatory biases. This will ultimately contribute to improved quality and effectiveness of human-robot interactions.

\bibliographystyle{IEEEtran}
\bibliography{sample-base}

\end{document}